%% file: main.tex
\documentclass[10pt,twocolumn,letterpaper]{article}

\usepackage[pagenumbers]{main}

\usepackage[table]{xcolor}
\usepackage{microtype}
\usepackage{graphicx}
\usepackage{booktabs}
\usepackage{amsmath}
\usepackage{amssymb}
\usepackage{mathtools}
\usepackage{amsthm}

\usepackage{color}
\usepackage{enumitem}
\usepackage{fontawesome5}
\usepackage{multirow}
\usepackage{makecell}

\usepackage{algorithmic}
\usepackage{algorithm}
\usepackage{etoolbox}

\definecolor{lblue}{RGB}{231, 66, 52}
\definecolor{tech1}{RGB}{0, 102, 204}      
\definecolor{tech2}{RGB}{220, 20, 60}      
\definecolor{tech3}{RGB}{34, 139, 34}      
\definecolor{tech4}{RGB}{255, 140, 0}      
\definecolor{tech5}{RGB}{138, 43, 226}     
\definecolor{tech6}{RGB}{184, 134, 11}     

\usepackage[pagebackref,breaklinks,colorlinks,citecolor=lblue]{hyperref}

\def\paperID{1}
\def\confName{RoboSense}
\def\confYear{IROS 2025}

\title{Enhancing Vision-Language Models for Autonomous Driving through Task-Specific Prompting and Spatial Reasoning}

\author{Aodi Wu$^{1,2}$, Xubo Luo$^{1,2}$\\
$^1$University of Chinese Academy of Sciences\\
$^2$Technology and Engineering Center for Space Utilization, Chinese Academy of Sciences\\
Beijing, China\\
{\tt\small wuaodi20@mails.ucas.ac.cn, luoxubo23@mails.ucas.ac.cn}
}

\begin{document}

\maketitle

\input{sections/0_abstract}

\input{sections/1_intro}
\input{sections/2_related_work}
\input{sections/3_method}
\input{sections/4_experiments}
\input{sections/5_conclusion}

{
\small
\bibliographystyle{ieeenat_fullname}
\bibliography{main}
}

\end{document}

%% file: sections/0_abstract.tex
\begin{abstract}
This technical report presents our solution for the RoboSense Challenge at IROS 2025, which evaluates Vision-Language Models (VLMs) on autonomous driving scene understanding across perception, prediction, planning, and corruption detection tasks. We propose a systematic framework built on four core components. First, a \emph{Mixture-of-Prompts router} classifies questions and dispatches them to task-specific expert prompts, eliminating interference across diverse question types. Second, \emph{task-specific prompts} embed explicit coordinate systems, spatial reasoning rules, role-playing, Chain-of-Thought/Tree-of-Thought reasoning, and few-shot examples tailored to each task. Third, a \emph{visual assembly} module composes multi-view images with object crops, magenta markers, and adaptive historical frames based on question requirements. Fourth, we configure \emph{model inference parameters} (temperature, top-$p$, message roles) per task to optimize output quality. Implemented on Qwen2.5-VL-72B, our approach achieves 70.87\% average accuracy on Phase-1 (clean data) and 72.85\% on Phase-2 (corrupted data), demonstrating that structured prompting and spatial grounding substantially enhance VLM performance on safety-critical autonomous driving tasks. Code and prompt are available at \url{https://github.com/wuaodi/UCAS-CSU-phase2}.
\end{abstract}

%% file: sections/1_intro.tex
\section{Introduction}

Autonomous driving systems require accurate scene understanding to make safe decisions. With the advancement of Vision-Language Models (VLMs)~\cite{openai2023gpt4v, liu2023llava, bai2025qwen2}, there is growing interest in applying these models to autonomous driving tasks. The RoboSense Challenge at IROS 2025 provides a comprehensive benchmark spanning perception, prediction, planning, and corruption detection. 

Despite their strong general capabilities, applying VLMs to autonomous driving faces three critical challenges. First, \emph{spatial reasoning in multi-view scenarios} is particularly difficult—VLMs often confuse left/right directions, misinterpret objects from BACK-view cameras as being ahead of the ego vehicle, and struggle with coordinate-based object references. Second, \emph{prompt interference across diverse task types} degrades performance when a single universal prompt attempts to cover perception, prediction, planning, and corruption detection simultaneously; optimizing the prompt for one task often hurts others. Third, \emph{temporal context integration} requires careful design—naively adding historical frames can introduce noise and distraction rather than helpful motion cues, especially when the temporal evidence is irrelevant to the specific question.

We address these challenges through three key design principles. First, we eliminate prompt interference by routing each question to a specialized expert prompt rather than using a single universal instruction. Second, we enhance spatial reasoning by explicitly defining multi-view coordinate systems and domain-specific constraints (e.g., BACK-camera objects are always behind the ego vehicle) that serve as grounding anchors. Third, we adapt temporal context per question type, selecting appropriate historical frame modes to provide relevant motion cues while avoiding distraction. These principles are realized through a four-component pipeline: a router, task-specific prompts with structured reasoning (Chain-of-Thought/Tree-of-Thought), adaptive visual assembly, and per-task inference parameters. Implemented on Qwen2.5-VL-72B, our system achieves 70.87\% on Phase-1 and 72.85\% on Phase-2.

%% file: sections/2_related_work.tex
\section{Related Work}
\label{sec:related}

\subsection{VLMs for Autonomous Driving}

Recent technical reports advance the general capability, reasoning and efficiency of open-source VLMs, such as Qwen2.5-VL~\cite{bai2025qwen2}, GLM-4.5V/4.1V-Thinking~\cite{vteam2025glm45vglm41vthinkingversatilemultimodal}, and InternVL3.5~\cite{wang2025internvl3}. These models provide strong backbones for driving-oriented perception and reasoning.

From the evaluation perspective, DriveBench~\cite{xie2025vlms} surfaces reliability issues highly relevant to safety-critical autonomy: weak visual grounding, sensitivity to corruptions, and multi-modal reasoning gaps. Ego3D-Bench~\cite{gholami2025spatial} further shows that multi-view, ego-centric spatial reasoning remains challenging for current VLMs and proposes Ego3D-VLM to inject structured spatial priors. Our method is complementary: instead of finetuning models, we target reliability by prompt-level design—explicit spatial grounding, adaptive temporal evidence, and task-specific routing—so that strong VLMs can be better aligned to autonomous-driving questions.

\subsection{Prompt Engineering}

Prompt engineering offers controllable ways to elicit reasoning and stabilize outputs. Role-playing provides a principled lens to steer behaviour~\cite{shanahan2023role}, and RoleLLM~\cite{wang2023rolellm} shows that explicit role profiles and role-conditioned tuning can enhance style control. Chain-of-Thought prompting elicits step-by-step reasoning~\cite{wei2022chain}; Tree-of-Thought extends this to explore multiple paths with self-evaluation~\cite{yao2023tree}; self-reflection further improves problem solving via post-hoc critique~\cite{renze2024self}. In-context learning~\cite{dong2022survey} shapes format and priors through compact exemplars, while visual prompting surveys techniques to ground attention on images~\cite{wu2024visual}. Beyond a single instruction, Mixture-of-Prompts~\cite{wang2024one} motivates using multiple expert prompts.

Guided by these insights, our system instantiates: (i) a router that dispatches queries to \emph{task-specific} prompts; (ii) an expert prompt design combining a coordinate-system block, reasoning guidance (CoT/ToT with brief self-check), and few-shot examples; and (iii) visual prompting with markers and crops plus adaptive history. This design leverages strong foundation VLMs while directly addressing reliability and spatial–temporal grounding concerns highlighted by prior studies.

%% file: sections/3_method.tex
\section{Methodology}
\label{sec:method}

\begin{figure*}[t]
\centering
\includegraphics[width=\textwidth]{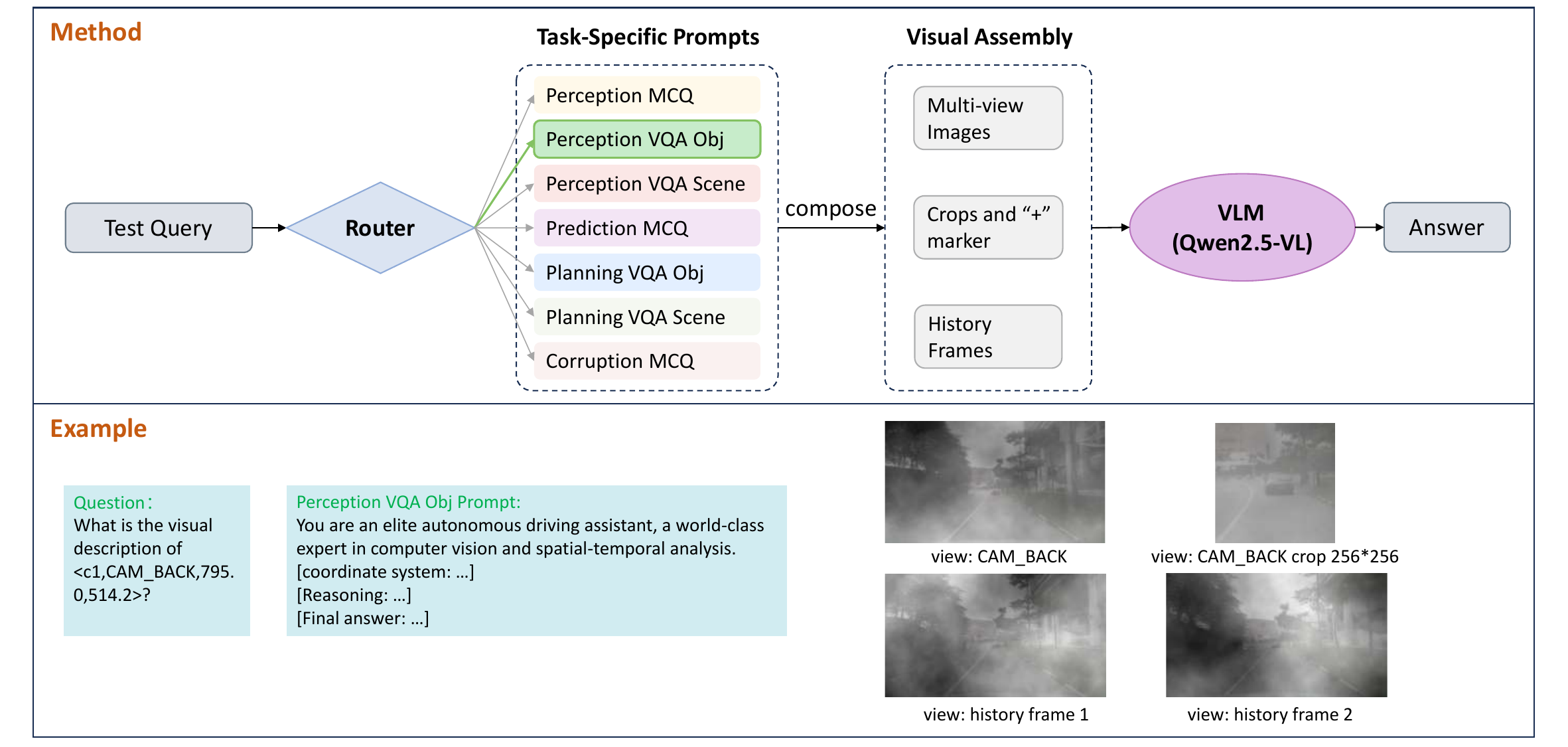}
\vspace{-2mm}
\caption{Method overview. A router classifies the test query and selects one task-specific expert (prompt) from multiple candidates. Each expert (shown in different colors) handles a specific question type with specialized instructions and examples. The activated expert's prompt is then combined with multi-view images, magenta "+" markers, region crops, and adaptive historical frames before invoking the VLM to produce the answer.}
\label{fig:overview}
\end{figure*}

Our approach leverages systematic prompt engineering to enhance VLM performance on autonomous driving scene understanding. As illustrated in Figure~\ref{fig:overview}, the pipeline consists of four core components: a \emph{Router} that classifies each test query and dispatches it to the appropriate task-specific expert; \emph{Task-Specific Prompts} that embed coordinate systems, spatial reasoning rules, role-playing, Chain-of-Thought/Tree-of-Thought reasoning, and few-shot examples tailored to each question type; a \emph{Visual Assembly} module that composes multi-view images, object crops with optional magenta markers, and adaptive historical frames; and finally the \emph{VLM} (Qwen2.5-VL) with task-dependent inference parameters that generates the answer. 

A key insight underlying all components is that explicit coordinate system definitions and spatial reasoning constraints are essential to ground VLMs in multi-view autonomous driving scenarios, where implicit scene understanding often fails. The following subsections detail each component in turn.

\subsection{Router}

A critical design choice in our system is to route questions to task-specific prompts~\cite{wang2024one} rather than using a single universal prompt. Early experiments revealed that when a unified prompt attempted to cover all question types with diverse instructions and examples, performance gains on some categories were often accompanied by drops on others. We attribute this to \emph{prompt confusion}: overly long or heterogeneous prompts can cause the model to struggle with conflicting guidance, leading to unstable behavior.

To mitigate this, we implement a lightweight router that classifies each query into one of seven supported types (Perception-MCQs, Perception-VQA-Object, Perception-VQA-Scene, Prediction-MCQs, Planning-VQA-Object, Planning-VQA-Scene, and Corruption-MCQs) and dispatches it to the corresponding specialized prompt. Given that the challenge questions follow fixed templates, we employ a rule-based approach leveraging the question category field provided in the dataset JSON and regular expressions to detect object references (e.g., \texttt{<c2,CAM\_VIEW,x,y>}). This simple strategy proves effective for the structured benchmark setting; for more open-ended or diverse question distributions, one could replace the router with a small LLM-based classifier.

Beyond prompt selection, the router also determines the visual assembly configuration (e.g., which history mode to use) and inference parameters, ensuring each question type receives tailored processing.

\subsection{Task-Specific Prompts}

Each question type is paired with a specialized prompt template. The templates share a compact and interpretable structure that are selected by the router at run time:

\paragraph{Coordinate system and spatial rules.}
We explicitly define the six-camera layout and the 1600$\times$900 pixel frame with origin at the top-left. Simple yet effective heuristics guide spatial grounding (e.g., \(x<800\) for left-half; \(y<450\) for upper region). Domain constraints further reduce errors: views from the BACK cameras are always behind the ego; motion is interpreted by orientation and road geometry; lane context hints moving vs. parked objects; and common driving conventions apply (e.g., double yellow lines as dividers, explicit evidence before declaring ``stopped'').

\paragraph{Role-playing.}
Each prompt assigns the model a task-specific expert role~\cite{shanahan2023role, wang2023rolellm} to guide its reasoning style and establish the appropriate analytical mindset. For instance, perception tasks cast the model as an ``elite autonomous driving assistant and world-class expert in computer vision and spatio-temporal analysis''; planning tasks assign it the role of a ``world-class autonomous driving risk analyst and strategic planning expert''; while prediction tasks position it as a specialist in ``motion forecasting and interaction prediction.'' This role-playing mechanism leverages the model's capacity for in-context persona adaptation, helping to elicit domain-relevant reasoning patterns and maintain consistency in output tone.

\paragraph{Chain-of-Thought reasoning.}
Instead of directly producing an answer, prompts instruct the model to reason step-by-step~\cite{wei2022chain}. The reasoning flow is structured as a sequence: (1) identify static properties from the current frame (object type, color, spatial position); (2) verify dynamic state by comparing position across historical frames to distinguish moving vs. parked objects; (3) synthesize spatial and temporal cues to form a hypothesis; and (4) perform a brief self-check to ensure consistency with domain rules (e.g., BACK-camera constraint, lane-context prior). For VQA tasks the reasoning is exposed in the output, providing interpretability; for MCQs it remains implicit but still guides the internal decision process.

\paragraph{Tree-of-Thought exploration.}
When the query admits multiple plausible outcomes (e.g., predicting whether an object will enter the ego's path), we apply a lightweight Tree-of-Thought strategy~\cite{yao2023tree}. The prompt explicitly asks the model to consider both branches and then determine which hypothesis is best supported by the visual evidence and motion trends. This dual-path exploration reduces premature commitment and improves robustness on ambiguous cases, albeit at modest additional computational cost.

\paragraph{Few-shot in-context examples.}
Each prompt includes 2--5 carefully curated examples~\cite{dong2022survey} that demonstrate the expected reasoning structure, evidence synthesis, and answer format. These examples show how to reference specific views (e.g., ``CAM\_BACK''), cite temporal changes (e.g., ``comparing T=0 and T-1 reveals...''), and integrate domain constraints. By anchoring the model's behavior to concrete demonstrations, we achieve more stable outputs and reduce formatting errors, particularly for complex multi-step questions.

\paragraph{Prompt example.}
Figure~\ref{fig:prompt_example} shows a condensed Perception-VQA-Object prompt template with key techniques highlighted in color.

\begin{figure}[h]
\centering
\fcolorbox{black!30}{gray!5}{
\begin{minipage}{0.95\columnwidth}
\small
\textcolor{blue}{\textbf{[Role-playing]}} You are an elite autonomous driving assistant, a world-class expert in computer vision and spatio-temporal analysis.

\vspace{2mm}
\textcolor{teal}{\textbf{[Coordinate System \& Spatial Rules]}} 

Six cameras: FRONT, FRONT\_LEFT/RIGHT, BACK, BACK\_LEFT/RIGHT. Image frame: 1600$\times$900, origin top-left.

Spatial heuristics: $x<800$ left-half, $x>800$ right-half; $y<450$ upper, $y>450$ lower.

Critical constraint: Any object in BACK cameras is ALWAYS behind the ego vehicle.

Motion rule: Determine parked vs. moving by comparing position across frames (T=0, T-1, T-2).

\vspace{2mm}
\textcolor{purple}{\textbf{[Chain-of-Thought Structure]}}

Your reasoning MUST follow: (1) Identify static properties from current frame \& crop; (2) Verify dynamic state via temporal comparison; (3) Synthesize evidence; (4) Self-check consistency with spatial rules.

\vspace{2mm}
\textcolor{orange}{\textbf{[Few-shot Example]}}

Question: What is \texttt{<c2,CAM\_BACK\_RIGHT,1324,552>}?

Reasoning: Object in CAM\_BACK\_RIGHT $\Rightarrow$ behind \& right of ego. Crop: dark sedan in parking area. Temporal check: position vs. fence unchanged across T=0/T-1/T-2 $\Rightarrow$ parked. Self-check: BACK constraint satisfied.

Final Answer: The object is a dark-colored sedan confirmed parked behind the ego vehicle.
\end{minipage}
}
\caption{Prompt template excerpt with techniques color-coded: \textcolor{blue}{role-playing}, \textcolor{teal}{coordinate grounding}, \textcolor{purple}{CoT structure}, \textcolor{orange}{in-context example}.}
\label{fig:prompt_example}
\end{figure}

\subsection{Visual Assembly}

Once the router selects a task-specific prompt, the system composes the visual inputs to accompany the textual query. This assembly adapts to the question type and the presence of object references, ensuring each query receives contextually appropriate evidence.

\paragraph{Multi-view images.}
All six camera views are transmitted in a canonical order (FRONT, FRONT\_RIGHT, FRONT\_LEFT, BACK, BACK\_RIGHT, BACK\_LEFT) to maintain consistent spatial correspondence across queries. For object-centric questions, only the views containing the referenced objects are sent to reduce noise; for scene-level questions, the full six-view set provides comprehensive spatial context.

\paragraph{Object-centric visual grounding.}
When a query references a specific object via coordinates (e.g., \texttt{<c2,CAM\_BACK\_RIGHT,x,y>}), we augment the full view with two additional cues~\cite{wu2024visual}. First, a magenta ``+'' marker is optionally overlaid at the object's location in the full image to facilitate visual alignment. Second, a 256$\times$256 crop centered at the coordinates is appended to provide fine-grained detail that may be lost in the downscaled full view. This dual representation—context plus detail—mirrors human strategies of alternating between scene overview and focused inspection.

\paragraph{Adaptive temporal context.}
Historical frames supply motion cues essential for prediction and planning. We support three modes selected by question type: \emph{grid} mode assembles all six views from each historical timestep into a labeled mosaic, ideal for scene-level reasoning; \emph{front} mode sends only the CAM\_FRONT view, concentrating on forward-driving context; and \emph{referenced} mode restricts history to the views containing the query's referenced object, minimizing irrelevant content. Typically, two historical frames (T-1 and T-2) precede the current frame, though corruption detection disables history to focus on single-frame artifact identification.

\begin{figure}[h]
\centering
\includegraphics[width=1.0\columnwidth]{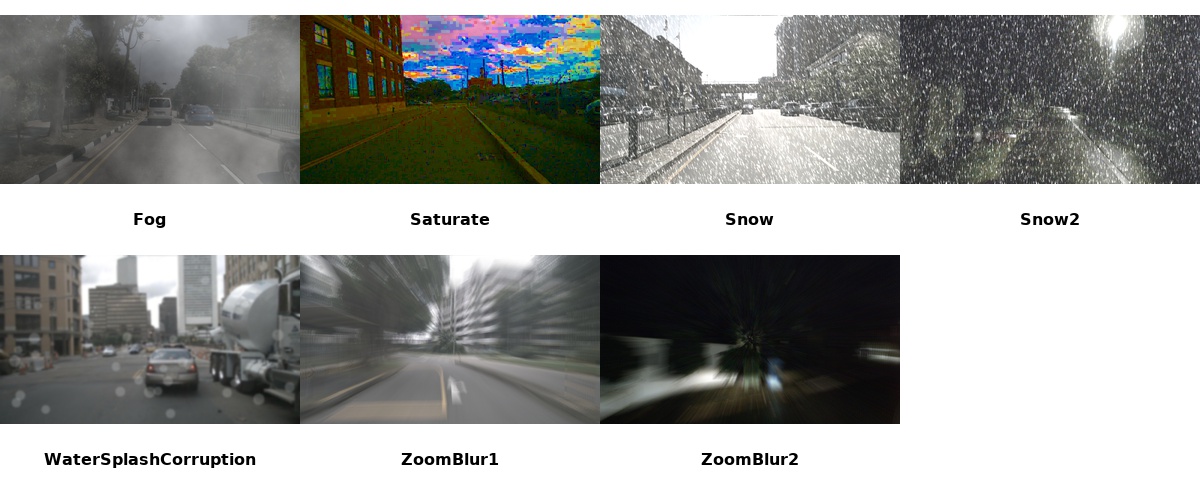}
\caption{The reference image used for corruption detection.}
\label{fig:corruption_ref}
\end{figure}

\paragraph{Corruption detection with reference images.}
For Corruption-MCQs, a reference image (Figure~\ref{fig:corruption_ref}) is prepended to the input. This reference image provides a visual baseline that helps the model distinguish normal appearance from degraded conditions such as fog, blur, or saturation.

\subsection{Model Selection and Inference Parameters}

We adopt Qwen2.5-VL-72B-Instruct as our primary backbone due to its strong performance on vision-language reasoning benchmarks and its availability as an open-source model. Beyond prompt engineering, we further tailor the model's behavior through task-specific inference configurations.

For tasks requiring deterministic and precise outputs—such as multiple-choice questions and object-level descriptions—we employ conservative decoding with low temperature (0.2) and low sampling threshold (top-$p$=0.2). This setting suppresses stochastic variation and yields stable, focused responses. Conversely, for scene-level descriptions where richness and contextual nuance are valued, we relax these constraints (temperature 1.5, top-$p$=0.9) to encourage broader exploration of the output space.

An additional refinement involves message role assignment. For most question types, the task-specific prompt is delivered as a \emph{system} message. However, we empirically found that Planning-VQA-Object achieves better performance when the prompt is sent as a \emph{user} message instead, likely due to improved output format compliance.

To support rapid experimentation, we implement a parallelized inference pipeline with fault-tolerant checkpointing. Samples are processed concurrently across multiple workers, with intermediate results saved atomically to prevent data loss. Selective update modes allow re-processing of specific question types or index ranges without re-running the entire dataset, accelerating the iteration cycle for prompt refinement.

%% file: sections/4_experiments.tex
\section{Experiments}
\label{sec:experiments}

\subsection{Experimental Setup}

\paragraph{Dataset and tasks.}
We evaluate on two competition phases with distinct datasets. \emph{Phase-1} uses clean images and covers six question types: Perception-MCQs, Perception-VQA-Object, Perception-VQA-Scene, Prediction-MCQs, Planning-VQA-Object, and Planning-VQA-Scene. \emph{Phase-2} uses corrupted images and adds a seventh type, Corruption-MCQs.

\paragraph{Implementation.}
Unless stated otherwise, we use Qwen2.5-VL-72B-Instruct with task-specific decoding parameters (Section~\ref{sec:method}). Inference runs through a remote API with parallel workers. All reported results are averaged scores across test samples in each question category.

\subsection{Ablation Studies}

We present ablation studies for both competition phases to demonstrate the incremental benefit of each design component. In Phase-1, we explore the design space on clean images, systematically building up the core framework from a naive baseline to a robust prompting system. Starting from a simple six-view concatenation, we progressively introduce crops with coordinate grounding, domain-specific rules, Chain-of-Thought reasoning, and finally the Mixture-of-Prompts router. In Phase-2, we inherit the MoP backbone established in Phase-1. The Phase-2 baseline achieves an average of 58.96—further refine each task-specific prompt, add adaptive temporal context, incorporate corruption-detection mechanisms, and tune model parameters. Tables~\ref{tab:phase1_ablation} and \ref{tab:phase2_ablation} show the progressive improvements across both phases, with colored entries highlighting which metrics each technique improves.

\subsubsection{Phase-1: Building the Prompting Framework}

\begin{table*}[t]
\centering
\scriptsize
\setlength{\tabcolsep}{4pt}
\caption{Phase-1 ablation study (clean data; 6 types). Each row adds one technique to the previous configuration. Colored entries show improvements brought by that technique.}
\begin{tabular}{lcccccccc}
\toprule
Configuration & P-MCQ & P-Obj & P-Scene & Pred-MCQ & Plan-Obj & Plan-Scene & Avg & $\Delta$ \\
\midrule
Baseline (six views + Q) & 56.60 & 38.59 & 69.92 & 59.77 & 58.44 & 49.83 & 53.79 & -- \\
\textcolor{tech1}{+ Crops \& coordinate system} & \textcolor{tech1}{64.15} & \textcolor{tech1}{46.89} & 63.12 & \textcolor{tech1}{64.37} & \textcolor{tech1}{68.86} & \textcolor{tech1}{60.56} & \textcolor{tech1}{60.84} & \textcolor{tech1}{+7.05} \\
\textcolor{tech2}{+ Domain rules/constraints} & \textcolor{tech2}{75.47} & 43.32 & 63.12 & \textcolor{tech2}{66.86} & 66.83 & \textcolor{tech2}{62.40} & \textcolor{tech2}{61.79} & \textcolor{tech2}{+0.95} \\
\textcolor{tech3}{+ CoT reasoning (reason-first)} & \textcolor{tech3}{100.00} & \textcolor{tech3}{57.43} & 62.81 & \textcolor{tech3}{72.99} & \textcolor{tech3}{79.79} & \textcolor{tech3}{64.83} & \textcolor{tech3}{69.62} & \textcolor{tech3}{+7.83} \\
\textcolor{tech4}{+ Mixture-of-Prompts (MoP)} & 100.00 & \textcolor{tech4}{64.47} & \textcolor{tech4}{76.17} & 72.99 & 75.21 & 64.64 & \textcolor{tech4}{70.87} & \textcolor{tech4}{+1.25} \\
\bottomrule
\end{tabular}
\label{tab:phase1_ablation}
\end{table*}

Key observations from Phase-1 ablation: (1) \emph{Visual grounding is critical}—adding crops and explicit coordinate system definitions yields the largest initial gain (+7.05 average), with particularly strong improvements on Planning-VQA-Object (+10.42) and Plan-Scene (+10.73), confirming that spatial understanding is a primary bottleneck for autonomous driving reasoning tasks. (2) \emph{Domain constraints prevent systematic errors}—while domain rules/constraints provide modest average gains (+0.95), they are essential for eliminating category-level mistakes such as describing BACK-camera objects as ``ahead'' or predicting unrealistic lane changes. This technique improves Perception-MCQs from 64.15 to 75.47. (3) \emph{CoT unlocks reasoning capability}—switching from result-first to reason-first delivers another substantial jump (+7.83), achieving perfect scores on Perception-MCQs (100.00) and lifting Planning-VQA-Object to 79.79. This validates the importance of explicit reasoning steps. (4) \emph{MoP eliminates prompt interference}—the final router-based Mixture-of-Prompts strategy addresses prompt confusion across diverse question types, notably boosting Perception-VQA-Object (+7.04) and Perception-VQA-Scene (+13.36), culminating in an overall average of 70.87.

\subsubsection{Phase-2: Task-Specific Refinements on Corrupted Data}

\begin{table*}[t]
\centering
\scriptsize
\setlength{\tabcolsep}{3pt}
\caption{Phase-2 ablation study (corruption data; 7 types). Each row adds one technique to the previous configuration. Colored entries show improvements brought by that technique.}
\begin{tabular}{lcccccccccc}
\toprule
Configuration & P-MCQ & P-Obj & P-Scene & Pred-MCQ & Plan-Scene & Plan-Obj & Corr-MCQ & Avg & $\Delta$ \\
\midrule
Baseline (MoP backbone) & 93.88 & 37.95 & 62.20 & 67.48 & 61.93 & 48.00 & 97.12 & 58.96 & -- \\
\textcolor{tech1}{+ Dynamic history mode} & \textcolor{tech1}{98.98} & \textcolor{tech1}{39.39} & \textcolor{tech1}{63.60} & \textcolor{tech1}{67.83} & \textcolor{tech1}{64.40} & \textcolor{tech1}{50.06} & 96.15 & \textcolor{tech1}{60.32} & \textcolor{tech1}{+1.36} \\
\textcolor{tech2}{+ Perception-Obj refinement} & 98.98 & \textcolor{tech2}{50.30} & \textcolor{tech2}{63.90} & 67.83 & \textcolor{tech2}{64.50} & \textcolor{tech2}{50.37} & 96.15 & \textcolor{tech2}{62.31} & \textcolor{tech2}{+1.99} \\
\textcolor{tech3}{+ Corruption reference image} & 98.98 & 50.29 & 63.71 & 67.83 & 64.32 & 50.12 & \textcolor{tech3}{100.00} & \textcolor{tech3}{62.38} & \textcolor{tech3}{+0.07} \\
\textcolor{tech4}{+ Tree-of-Thought (ToT)} & \textcolor{tech4}{100.00} & 50.23 & \textcolor{tech4}{63.80} & \textcolor{tech4}{95.47} & 63.00 & 49.37 & 100.00 & \textcolor{tech4}{71.64} & \textcolor{tech4}{+9.26} \\
\textcolor{tech5}{+ Task-specific decoding params} & 100.00 & 49.92 & \textcolor{tech5}{64.24} & 95.47 & \textcolor{tech5}{64.12} & \textcolor{tech5}{49.90} & 100.00 & \textcolor{tech5}{72.01} & \textcolor{tech5}{+0.37} \\
\textcolor{tech6}{+ Planning-Obj as user role} & 100.00 & \textcolor{tech6}{49.97} & \textcolor{tech6}{64.51} & 95.47 & \textcolor{tech6}{64.20} & \textcolor{tech6}{53.10} & 100.00 & \textcolor{tech6}{72.85} & \textcolor{tech6}{+0.84} \\
\bottomrule
\end{tabular}
\label{tab:phase2_ablation}
\end{table*}

Notable findings from Phase-2 ablation: (1) \emph{Dynamic history mode provides temporal context}—adaptively selecting historical frames based on question type (grid for general awareness, front-only for motion prediction, and referenced for specific inquiries) improves multiple categories, with the average rising from 58.96 to 60.32 (+1.36). This indicates that temporal information helps, but must be task-aligned to avoid distraction. (2) \emph{Task-specific prompt refinement is highly effective}—individually optimizing the Perception-VQA-Object prompt (adding stricter directional constraints and implicit reasoning) brings that category from 37.95 to 50.30, a +12.35 gain, demonstrating the value of tailored instructions. (3) \emph{Corruption detection benefits from reference images}—adding the clean reference image for Corruption-MCQs achieves perfect scores (100.00), validating our visual assembly strategy for corruption-aware tasks. (4) \emph{ToT yields the largest improvement}—incorporating Tree-of-Thought reasoning (exploring multiple hypotheses before conclusion) produces a dramatic +9.26 gain, especially on Prediction-MCQs (67.48 $\to$ 95.47), where uncertainty is high. (5) \emph{Parameter tuning and role assignment refine outputs}—task-specific temperature/top-$p$ values and switching Planning-VQA-Object prompts to user-role further improve format compliance, bringing the final average to 72.85.

\subsection{Model Comparison}

We compare Qwen2.5-VL-72B with other state-of-the-art VLMs on both competition phases. Table~\ref{tab:model_comparison} reports results under identical prompting conditions.

\begin{table*}[t]
\centering
\scriptsize
\setlength{\tabcolsep}{4pt}
\caption{Model comparison across both phases under identical prompts and inference settings; best per column in \textbf{bold}.}
\begin{tabular}{llccccccccc}
\toprule
Phase & Model & P-MCQ & P-Obj & P-Scene & Pred-MCQ & Plan-Scene & Plan-Obj & Corr-MCQ & Avg \\
\midrule
\multirow{2}{*}{Phase-1} & Qwen2.5-VL-72B & \textbf{64.15} & 46.89 & \textbf{63.12} & \textbf{64.37} & \textbf{60.56} & \textbf{68.86} & -- & \textbf{60.84} \\
& GLM-4.5V & 37.74 & \textbf{51.48} & 53.98 & 57.85 & 52.15 & 37.34 & -- & 51.93 \\
\midrule
\multirow{2}{*}{Phase-2} & Qwen2.5-VL-72B & 93.88 & \textbf{37.95} & \textbf{62.20} & \textbf{67.48} & \textbf{61.93} & \textbf{48.00} & \textbf{97.12} & \textbf{58.96} \\
& InternVL-3.5 & \textbf{94.90} & 34.26 & 60.41 & 61.79 & 55.43 & 43.67 & 91.35 & 54.31 \\
\bottomrule
\end{tabular}
\label{tab:model_comparison}
\end{table*}

In Phase-1, while GLM-4.5V excels on Perception-VQA-Object (51.48 vs. 46.89), Qwen2.5-VL shows stronger overall consistency and significantly outperforms on Planning-VQA-Object (68.86 vs. 37.34). In Phase-2, Qwen2.5-VL-72B continues to outperform InternVL-3.5 on average (58.96 vs. 54.31), with particularly strong advantages on Perception-VQA-Object and Corruption-MCQs. We therefore adopt Qwen2.5-VL as our primary backbone.

%% file: sections/5_conclusion.tex
\section{Conclusion}
\label{sec:conclusion}

This technical report presents a systematic approach to autonomous driving scene understanding using Vision-Language Models. We address the challenges of the RoboSense Challenge through task-specific prompt engineering, spatial reasoning enhancement, and adaptive visual assembly.

\subsection{Key Contributions and Insights}

This work presents three main contributions for autonomous driving scene understanding with VLMs. First, we introduce a \emph{Mixture-of-Prompts} router that classifies questions and dispatches task-specific prompts to eliminate interference across diverse question types. Second, we establish \emph{coordinate grounding} through explicit camera layouts, spatial rules (e.g., BACK-camera constraints), and visual markers that significantly improve multi-view reasoning. Third, we develop \emph{structured reasoning strategies} including Chain-of-Thought for step-by-step analysis and Tree-of-Thought for exploring alternative hypotheses in uncertain scenarios. Our experiments reveal three critical insights: spatial understanding remains a bottleneck for VLMs but can be substantially improved through explicit coordinate systems and domain constraints; reasoning-first approaches consistently outperform direct-answer strategies; and task-specific prompt optimization is more effective than generic instructions when combined with appropriate visual and temporal context.

\subsection{Limitations and Future Work}

Our approach has three main limitations. First, autonomous driving is inherently a temporal task, yet our current implementation uses only two historical frames; longer video sequences may be necessary to capture motion trends and derive better driving strategies. Second, Chain-of-Thought and Tree-of-Thought reasoning significantly increase inference costs, limiting practical deployment. Third, spatial understanding challenges—particularly in multi-view geometry—may require solutions at the model training or fine-tuning stage rather than relying solely on prompt engineering.

Future work should explore long-term temporal modeling with extended video contexts, efficient reasoning mechanisms to reduce computational overhead, and VLM fine-tuning with autonomous driving data to fundamentally enhance spatial reasoning capabilities. Our results demonstrate that structured prompting substantially improves VLM performance on autonomous driving tasks, providing a foundation for these future directions.

\section*{Acknowledgments}

We thank the organizers of the RoboSense Challenge for providing this comprehensive benchmark and the opportunity to advance research in autonomous driving scene understanding.